\documentclass{article}

\usepackage[final]{corl_2020} 

\usepackage{graphicx}
\usepackage{wrapfig}

\usepackage{amsmath}
\usepackage{amsthm}
\usepackage{amsfonts}
\usepackage{stmaryrd}
\usepackage{algorithm}
\usepackage[noend]{algpseudocode}

\newboolean{showappendix}
\setboolean{showappendix}{true}
\newboolean{reftoappendix}
\setboolean{reftoappendix}{true}

\newcommand{\bc}[1]{\left\{{#1}\right\}}
\newcommand{\br}[1]{\left({#1}\right)}
\newcommand{\bs}[1]{\left[{#1}\right]}
\newcommand{\abs}[1]{\left| {#1} \right|}
\newcommand{\ip}[2]{\left\langle{#1},{#2}\right\rangle}
\newcommand{\norm}[1]{\left\| {#1} \right\|}

\newcommand{\E}[1]{\mathbb{E}\bs{{#1}}}

\DeclareMathOperator*{\argmax}{arg\,max}

\title{Interaction-limited Inverse Reinforcement Learning}

\author{%
  Martin Troussard \\
  Idiap Research Institute and EPFL \\
  \texttt{martin.troussard@idiap.ch} \\
  \And 
  Emmanuel Pignat \\
  Idiap Research Institute and EPFL \\
  \texttt{emmanuel.pignat@idiap.ch} \\
  \And
  Parameswaran Kamalaruban \\ 
  LIONS, EPFL \\ 
  \texttt{kamalaruban.parameswaran@epfl.ch} \\
  \And
  Sylvain Calinon \\
  Idiap Research Institute \\
  \texttt{sylvain.calinon@idiap.ch} \\
  \And
  Volkan Cevher \\ 
  LIONS, EPFL \\ 
  \texttt{volkan.cevher@epfl.ch} \\
}

\begin{document}
\maketitle


\begin{abstract}
This paper proposes an inverse reinforcement learning (IRL) framework to accelerate learning when the learner-teacher \textit{interaction} is \textit{limited} during training. Our setting is motivated by the realistic scenarios where a helpful teacher is not available or when the teacher cannot access the learning dynamics of the student. We present two different training strategies: Curriculum Inverse Reinforcement Learning (CIRL) covering the teacher's perspective, and Self-Paced Inverse Reinforcement Learning (SPIRL) focusing on the learner's perspective. Using experiments in simulations and experiments with a real robot learning a task from a human demonstrator, we show that our training strategies can allow a faster training than a random teacher for CIRL and than a batch learner for SPIRL.
\end{abstract}


\section{Introduction}\label{sec:intro}

Learning from Demonstrations (LfD) is an active research area that addresses the problem of learning how to perform a task by observing the demonstrations provided by an expert. This approach plays an important role in many real-life learning settings, including human-to-robot interaction~\cite{schaal1997learning,Billard08chapter,argall2009survey,chernova2014robot,kober2013reinforcement}. 

The two popular approaches for LfD include (i) behavioral cloning, which directly mimics the expert behavior, without understanding the objective \cite{bain1999framework}, and (ii) inverse reinforcement learning (IRL), which infers the reward function (i.e., the objective of the task) explaining the expert behavior \cite{russell1998learning}. 

In this work, we focus on the IRL approach to LfD. Typically, the IRL learner assumes that the demonstrated expert behavior is optimal with respect to some reward function, even if the reward function cannot be specified explicitly as in typical reinforcement learning (RL). IRL algorithms operate by first recovering this reward function from demonstrations, and then obtaining a  policy corresponding to the inferred reward~\cite{russell1998learning,abbeel2004apprenticeship}.
IRL has been extensively studied in the context of designing efficient learning algorithms for a given set of demonstrations \cite{abbeel2004apprenticeship,ratliff2006maximum,ziebart2008maximum,boularias2011relative}.

In this paper, we consider a sequential variant of the popular IRL algorithm, namely Maximum Entropy IRL algorithm \cite{ziebart2008maximum,rhinehart2017first}, where at each time step, the learner receives only a minibatch of demonstrations from the teacher. Recent works in this field have shown that the interaction between the learner and the teacher can considerably improve the speed of the learning process. 

From the learner's point of view, \cite{amin2017repeated,dorsa2017active} have proposed active learning algorithms for IRL that focus on reducing the number of demonstrations that needs to be requested from a teacher. From the viewpoint of a teacher, \cite{kamalaruban2019interactive} have designed active teaching algorithms for IRL, where the teacher actively chooses the appropriate next demonstration for the learner based on the learner's current progress and the target knowledge. 

However, to our knowledge, no prior work considers the problem of speeding up the learning process of an online IRL learner when the teacher-learner interaction during the learning process is not possible. To solve this problem, we extend some of the techniques that were initially developed to accelerate the training process in the supervised learning problem, to the problem of online IRL with the goal of minimizing the number of demonstrations required to learn/teach a task.

To understand the key contributions, consider the example of a teacher who wants to write a mathematics book consisting of some concepts related to a particular topic. The teacher could present the contents (definitions and theorems etc.) in a random order, hoping that the readers would find their path, or the teacher could organize the materials in a manner that is beneficial to an ``average'' reader, usually from more straightforward concepts to complex ones.  

This cognitive/learning process of humans is the intuition behind Curriculum Learning \cite{bengio2009curriculum} whose method consists in ordering the input data such that the ordering can improve the robustness and speed of the learning process of an online learner. Now, consider a reader who receives this book to learn some concepts. She does not need to follow the path given by the author exactly; for example, she might skip some parts, go back on others, etc. This insight is the basis for self-paced learning \cite{kumar2010self}, a learning algorithm in which the learner actively chooses the examples that she will process at each time step, i.e., the learner is designing her own curriculum.

\begin{wrapfigure}{r}{0.6\textwidth}
  \begin{center}
    \includegraphics[width=0.58\columnwidth]{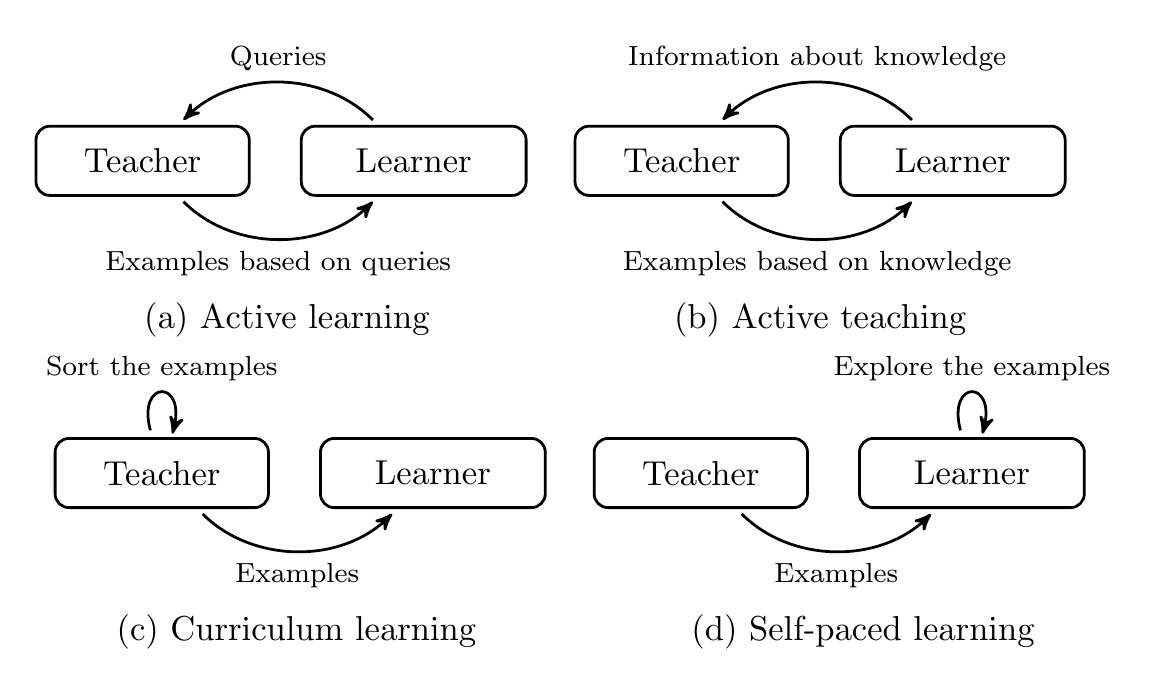}
  \end{center}
  \caption{Summary of different learning modalities}
  \label{fig:diagrams}
\end{wrapfigure}

Curriculum learning and self-paced learning have been applied effectively to the supervised and semi-supervised learning problems~\cite{bengio2009curriculum,zaremba2014learning,bengio2015scheduled,graves2017automated,jiang2018mentornet,ma2017self}. More recently, these techniques have been integrated with reinforcement learning algorithms as well~\cite{florensa2017reverse,svetlik2017automatic,narvekar2018learning}. Recent work \cite{jiang2015self} combines curriculum learning and self-paced learning to extract the benefits of both a careful teacher through the curriculum and a conscientious learner through self-paced learning. The different learning/teaching modalities presented in this section, and their similarities, are summarized in Figure~\ref{fig:diagrams}.


In this paper, we introduce Curriculum Inverse Reinforcement Learning (CIRL), a teaching strategy, and Self-Paced Inverse Reinforcement Learning (SPIRL), a learning strategy, which improve the learning speed of an agent that iteratively learns a task from demonstrations. To test our proposed algorithms, experimental results demonstrate the superiority of the proposed strategies in gridworld environments and with a real robot as compared to uninformative training strategies.

\section{Related Work}\label{sec:relatedwork}
Interaction between teacher and learner plays an essential role in accelerating the learning process of an IRL learner. \cite{lopes2009active} extends the principles of active learning to inverse reinforcement learning by allowing the agent to query the demonstrator for samples at specific states instead of receiving demonstrations at random. In a recent work \cite{amin2017repeated}, the learner has to navigate actively into a sequence of tasks demonstrated by a teacher, to minimize the number of queries made to the teacher. 

\cite{kamalaruban2019interactive} takes the viewpoint of a teacher in how to best assist the learning agent by providing an optimal sequence of demonstrations, using the learner's knowledge about the task. \cite{hadfield2016cooperative} have studied the interaction between the teacher and the learner as a cooperative game and have produced behaviors similar to active learning and active teaching. However, contrary to our approach, all these works suppose that teacher-learner interaction is possible during the whole learning process.

In \cite{cakmak2012algorithmic}, authors have studied the problem of improving the learning process of an IRL agent \cite{ng2000algorithms} in the batch setting by selecting the most informative set of demonstrations at once. In recent work, \cite{brown2018machine} have shown that the problem of finding the optimal set of demonstrations can be formulated as a set cover problem, and proposed machine teaching algorithms \cite{DBLP:journals/corr/ZhuSingla18,liu2017iterative,goldman1995complexity} for the batch IRL learner. 

Similar to our approach, there is no teacher-learner interaction during the learning process in these approaches as well, and the agents have to deal with the absence of feedback related to the learner's current knowledge. However, our approach focuses on the ways a fixed batch of demonstrations can be ordered by the teacher (in CIRL) or by the learner (in SPIRL), whereas machine teaching focuses on how this batch of demonstrations can be built.

\section{Background}\label{sec.model}

This section provides the necessary background on Inverse Reinforcement Learning, and learning modalities. 

\subsection{Inverse Reinforcement Learning (IRL)}

Consider an environment defined as a Markov Decision Process (MDP) $\mathcal{M} := \br{\mathcal{S},\mathcal{A},T,\gamma,P_0,R}$. The sets of possible states and actions are denoted by $\mathcal{S}$ and $\mathcal{A}$ respectively. $T: \mathcal{S} \times \mathcal{S} \times \mathcal{A} \rightarrow \bs{0,1}$ captures the state transition dynamics, i.e., $T\br{s' \mid s,a}$ denotes the probability of landing in state $s'$ by taking action $a$ from state $s$. Here $\gamma$ is the discounting factor, and $P_0: \mathcal{S} \rightarrow \bs{0,1}$ is an initial distribution over states $\mathcal{S}$. Given a feature mapping $\phi: \mathcal{S} \times \mathcal{A} \rightarrow \mathbb{R}^{d}$, each state-action pair $\br{s,a}$ is encoded by a feature vector $\phi\br{s,a} \in \mathbb{R}^d$. The reward function $R: \mathcal{S} \times \mathcal{A} \rightarrow \mathbb{R}$ is represented as a parameterized linear function with $w \in \mathbb{R}^d$ as $R_w\br{s,a} = \ip{w}{\phi\br{s,a}}$.

We denote a policy $\pi: \mathcal{S} \times \mathcal{A} \rightarrow \bs{0,1}$ as a mapping from a state to a probability distribution over actions. For any policy $\pi$, the value (total expected reward) of $\pi$ in the MDP $\mathcal{M}$ is defined as follows:
\[
V_\pi ~:=~ \E{\sum_{t=0}^{\infty} \gamma^t R_w(s_t, a_t)~\bigg\vert~s_0 \sim P_{0}, \pi}
~=~ w^\top \E{\sum_{t=0}^{\infty} \gamma^t \phi(s_t, a_t)~\bigg\vert~s_0 \sim P_{0}, \pi} ~=~ \ip{w}{\mu_\pi} ,
\]
where $\mu_\pi = \E{\sum_{t=0}^{\infty} \gamma^t \phi(s_t, a_t)~\bigg\vert~s_0 \sim P_{0}, \pi}$ is the feature expectation vector of the policy $\pi$. Similarly, for any trajectory $\xi = \bc{\br{s_t,a_t}}_{t = 0,1,\dots}$, representing a sequence of state-action pairs, we define the empirical feature expectation vector as $\mu_\xi := \sum_{t=0}^{\infty} \gamma^t \phi(s_t, a_t)$. Then for a set of trajectories $\Xi$, we have $\mu_\Xi = \frac{1}{\abs{\Xi}} \sum_{\xi \in \Xi} \mu_\xi$.

For a fully specified MDP $\mathcal{M}$ (with reward function $R_{w^*}$ where $\norm{w^*}_1 \leq 1$), an optimal policy $\pi^*$ is given by $\pi^* = \argmax_\pi V_\pi$. We refer $\pi^*$ as the expert policy that we want to teach to the learner. In IRL, the learner does not know the reward function $R_{w^*}$, and must infer it from the set of trajectories $\Xi$ given by the expert. Each trajectory is obtained by executing the policy $\pi^*$ in the MDP $\mathcal{M}$. The performance of a learned policy $\pi$ (w.r.t. $\pi^*$) in $\mathcal{M}$ is evaluated via the feature expectation mismatch given by $\norm{\mu_\pi - \mu_{\pi^*}}_2 $ \cite{abbeel2004apprenticeship}.

\subsection{Maximum Entropy IRL (MaxEnt IRL)}

\begin{wrapfigure}{R}{0.5\textwidth}
    \begin{minipage}{0.5\textwidth}
      \begin{algorithm}[H]
        \caption{Online MaxEnt IRL}\label{algo:onlinemaxent}
        \begin{algorithmic}
        \State Initialize $w_0$
	\For{$t=0, \ldots, T$}
	\State Compute $\pi_{w_t}$ via Soft Value Iteration
	\State Receive a batch of demonstrations $\Xi_t$
	\State $w_{t+1} \gets w_{t} + \eta_t \br{\mu_{\Xi_t} - \mu_{\pi_{w_t}}}$
	\EndFor
	\State \Return $w_T$
        \end{algorithmic}
      \end{algorithm}
    \end{minipage}
  \end{wrapfigure}

In the MaxEnt IRL \cite{ziebart2008maximum,ziebart2010modeling} formulation, the learner seeks to maximize the log-likelihood 
\begin{align*}
\mathcal{L}(\Xi, w) ~=~& \sum_{\xi \in \Xi} \log \mathbb{P}(\xi | w) \\
~=~ &\sum_{\xi \in \Xi} \sum_{t=0}^{\infty} \log \pi_w(a^{(\xi)}_t|s^{(\xi)}_t) ,
\end{align*}
where $\pi_w$ is the policy obtained by Soft Value Iteration \cite[Algorithm.~9.1]{ziebart2010modeling} with parameter $w$\ifthenelse{\boolean{reftoappendix}}{ (see Appendix \ref{appendix:maxentirl})}{}. Note that the function $\mathcal{L}$ is concave, and its gradient is given by
\[
\frac{\partial\mathcal{L}}{\partial w}(\Xi, w) ~=~ \mu_\Xi - \mu_{\pi_w}.
\]
Given a batch of expert demonstrations $\Xi$, the optimal reward weights vector $w^*$ can be obtained by gradient ascent on $\mathcal{L}$. In this work, we consider a natural online variant (cf. \cite{rhinehart2017first}) of the batch MaxEnt IRL. In this setting, at each time step $t$, the learner receives a mini-batch of demonstrations $\Xi_t$ and perform a gradient ascent step. Here, the gradient will be $g = \mu_{\Xi_t} - \mu_{\pi_{w_t}}$, the difference between the feature expectation vector of $\Xi_t$ and the learner's current feature expectation vector. Algorithm~\ref{algo:onlinemaxent} presents the online MaxEnt IRL learner. 

  


\subsection{Curriculum and Self-Paced Learning}

The field of curriculum learning attempts to rank the training examples from the ``easy'' to ``hard''  ones based on a ``difficulty score''. Specifically, we find an ordering $\gamma : \mathcal{D} \to \mathbb{N}$ of the training examples $\mathcal{D} = \{(x_i, y_i)\}_{1 \leq i \leq n}$ such that an online learner (for example a learner using Stochastic Gradient Descent) is exposed to the training data in this particular order. There are many ways to define a curriculum, often depending on the problem of interest. 

In \cite{bengio2009curriculum}, the authors rank images of geometrical shapes according to their variability (for example circles are ranked before ellipsoids). In \cite{weinshall2018theory}, the authors suggest to use the loss of the optimal hypothesis as a difficulty score and prove speedup in convergence for linear regression and binary classification (with hinge loss as the error measure).

Even though designing a generic curriculum for different types of learners can be intuitively good, this approach is limited by the fact that the teacher is not taking into account the current progress of the learner. Self-paced learning \cite{kumar2010self} addresses this limitation by allowing the curriculum to be designed by the learner. In particular, given a loss function $L(y_i, f_w (x_i))$ between the ground truth data $y_i$ and the estimated output $f_w (x_i)$, the learner seeks to minimize
\[
\mathfrak{L}(w, v ; \lambda) ~=~ \sum_{i=1}^n v_i L(y_i, f_w (x_i)) - \lambda \sum_{i=1}^n v_i,
\]
where $v \in [0,1]^n$ is a weight vector specifying the importance given to each training example. 

\begin{wrapfigure}{R}{0.5\textwidth}
    \begin{minipage}{0.5\textwidth}
      \begin{algorithm}[H]
        \caption{Self-Paced Learning}\label{algo:spl}
        \begin{algorithmic}
        \State Input: Training examples $\mathcal{D} = \{(x_i, y_i)\}_{1 \leq i \leq n}$
	\State Initialize $w_0$
	\For{$t=0, \ldots, T-1$}
	\State $\mathcal{D}_t = \{(x, y) \in \mathcal{D}~|~L(y, f_{w_{t}} (x)) \leq \lambda\}$
	\State $w_{t+1} = \underset{w}{\arg\min}~\sum_{(x,y) \in \mathcal{D}_t} L(y, f_w (x))$
	\If{$\lambda$ is too small} increase $\lambda$ \EndIf
	\EndFor
	\State \Return $w_T$
        \end{algorithmic}
      \end{algorithm}
    \end{minipage}
  \end{wrapfigure}

This objective function is biconvex and can be optimized by alternating minimization of $\mathfrak{L}$ with respect to $w$ and $v$. With the fixed $w$, we note that $v^* = \underset{v}{\arg\min}~\mathfrak{L}(w, v ; \lambda)$ is given by
\[
v_i^* ~=~ \begin{cases}
1 & \text{if } L(y_i, f_w (x_i)) < \lambda,\\
0 & \text{otherwise}.
\end{cases}
\]
This means that at each iteration of the learning process, the learner will select only training examples whose loss is below the threshold $\lambda$. Intuitively, the parameter $\lambda$ should increase over time to be able to select more and more difficult examples with time (a basic rule is for example to increase $\lambda$ when the set of training examples considered is not growing). The overall procedure is described in Algorithm~\ref{algo:spl}.


\section{Interaction-limited IRL}\label{sec.main.result}

In this section, we derive two strategies, one for teaching and the other for learning, that seek to improve the training process of an online IRL learner when the learner can not interact with the teacher during the learning process.

\paragraph{Theoretical Insights}

We define the loss function of a trajectory $\xi$ under the reward weight vector $w$ as
\[
\ell(\xi, w) ~=~ - \log \mathbb{P}(\xi | w).
\]
Consider a learner using the Online MaxEnt IRL, as presented in Algorithm~\ref{algo:onlinemaxent}. At a fixed time $t$ of the learning process, using the update rule $w_{t+1} = w_{t} + \eta_t (\mu_{\Xi_t} - \mu_{\pi_{w_t}})$, we have
\[
\norm{w_{t+1} - w^*}^2 ~=~ \norm{w_t - w^*}^2 + \eta_t^2 \norm{\mu_{\Xi_t} - \mu_{\pi_{w_t}}}^2  
+ 2 \eta_t \ip{w_t - w^*}{\mu_{\Xi_t} - \mu_{\pi_{w_t}}} .
\]
The objective of a greedy teacher would be to provide a batch of demonstrations $\Xi_t$ such that
\[
\Xi_t ~=~ \underset{\Xi'}{\arg \min}~~ \norm{w_{t+1} - w^*}^2 
~=~  \underset{\Xi'}{\arg \min}~~2 \ip{w_t - w^*}{\mu_{\Xi'}} + \eta_t \norm{\mu_{\Xi'} - \mu_{\pi_{w_t}}}^2 .
\]

\paragraph{Teacher:} If the teacher has to choose the next batch, without knowing the current knowledge of the learner (i.e., she only has access to $w^*$), this teacher could seek to find $\underset{\Xi'}{\arg \min}~~- \ip{w^*}{\mu_{\Xi'}}$, which corresponds to the demonstrations with high reward with respect to $w^*$. This observation matches with that of \cite{weinshall2018theory} regarding curriculum learning in the context of supervised learning problem.

\paragraph{Learner:} Conversely, if the learner has to choose the next batch, without the possibility of asking to a teacher (i.e., she only has access to $w_t$), this learner would like to find $\underset{\Xi_t}{\min}~~2 \ip{w_t}{\mu_{\Xi_t}} + \eta_t \norm{\mu_{\Xi_t} - \mu_{\pi_{w_t}}}^2$. This objective can be interpreted as follows: 
\begin{itemize}
\item In the beginning of the learning process, $\eta_t$ is large, thus the learner will give more attention to minimize $\| \mu_{\Xi_t} - \mu_{\pi_{w_t}} \|^2$, which means she will select a demonstration whose features expectation vector is close to $ \mu_{\pi_{w_t}}$, which also means that she will choose demonstrations with high reward with respect to $w_t$.
\item After a sufficient number of iterations, $\eta_t$ is close to $0$, the learner will minimize $\langle w_t , \mu_{\Xi_t} \rangle$, which means the demonstrations with low reward with respect to $w_t$, which can be viewed as ``outliers.''
\end{itemize}
This observation matches with the intuition behind Self-Paced Learning.

\subsection{Curriculum Inverse Reinforcement Learning}
\label{subsection:CIRL}

In this section, we extend the principles of curriculum learning to Inverse Reinforcement Learning. Following the insights proposed in \cite{weinshall2018theory}, we rank the demonstrations of $\Xi$ according to their loss with respect to the true reward weight vector $w^*$, i.e., $\ell(\xi, w^*)$. There are two possible interpretations of this objective function:
\begin{itemize}
\item \textbf{Reward-based curriculum (R-CIRL)}: Using the fact that, in the MaxEnt IRL framework, $$\mathbb{P}(\xi | w^*) \propto \exp\left\langle w^*, \mu_\xi\right\rangle,$$ a first curriculum consists of ordering the trajectories of $\Xi$ according to their reward (under the true reward weight vector $w^*$).
\item \textbf{Probability-based curriculum (P-CIRL)}: Another way of ranking the demonstrations would be to directly compute, for a trajectory $\xi$ of finite length $H$, $\mathbb{P}(\xi | w^*)$ based on the state-action probability 
\[
\mathbb{P}(\xi | w^*) = \prod_{t=1}^H \pi_{w^*}(a^{(\xi)}_t|s^{(\xi)}_t).
\]
\end{itemize}

\subsection{Self-Paced Inverse Reinforcement Learning}


In this section, we use $\ell(\xi, w)$ as a loss function to extend the principles of self-paced learning to IRL. The resulting procedure, presented in Algorithm~\ref{algo:spirl}, is a special case of the online MaxEnt IRL algorithm presented in Algorithm~\ref{algo:onlinemaxent}.

\begin{algorithm}
\caption{Self-Paced Inverse Reinforcement Learning}\label{algo:spirl}
\begin{algorithmic}[1]
\Require{Expert demonstrations $\Xi = \{\xi_i\}_{1 \leq i \leq n}$}
\State Initialize $w_0$
\For{$t=0, \ldots, T-1$}
\State $\pi_{w_t}$ is obtained by Soft Value Iteration
\State $\Xi_t = \Big\{\xi \in \mathcal{D}~\Big|~- \sum_{t=0}^{\infty} \gamma^t \log \pi_{w_t}(a^{(\xi)}_t|s^{(\xi)}_t) \leq \lambda\Big\}$
\State $w_{t+1} = w_{t} + \eta_t (\mu_{\Xi_t} - \mu_{\pi_{w_t}})$
\If{$\lambda$ is too small} increase $\lambda$ \EndIf
\EndFor
\State \Return $w_T$
\end{algorithmic}
\end{algorithm}

\section{Experiments}\label{sec:experiments}

In this section, we demonstrate the performance of our algorithms in gridworld environments. In Appendix~\ref{appendix:robotics}, we provide results for experiments with a real robot. 

\subsection{Experimental Setup}

\paragraph{Preliminaries} The error metric used in our experiments is the distance between the learner's and the teacher's features expectation vectors. Because we want to study the effect of the demonstrations' ordering on the learning process, when a demonstration is provided, it is discarded from the pool.

\paragraph{Baselines} CIRL focuses on the ordering of a demonstrations' set from the teacher point of view for an online learner, thus we compare CIRL to a teacher giving demonstrations at random (i.e. a teacher with no particular strategy) and a teacher doing the opposite of what the curriculum suggests (anticurriculum). SPIRL focuses on how to process actively, from the learner's point of view, a fixed batch of demonstrations (the learner has access to the whole batch at any time), we compare it to a learner computing his gradient with the full batch of demonstrations at each time step.

\paragraph{Learner Model} The learner is an online learner, as described in Algorithm \ref{algo:onlinemaxent}, using gradient ascent with learning rate $\eta_t = \frac{1}{t}$.

\paragraph{Learner Initialization} Since the quality of the sequence of demonstrations provided highly depends on the initial $w$ of the learner, we will repeat the process several times (around 200 times for each environment and each learning/teaching modality) with a different initialization of the learner's reward weights $w_0$, sampled uniformly in $[-1,1]^d$.

\subsection{Discrete environments}
\label{subsection:discrete}

\begin{figure}[ht]
\begin{minipage}[b]{0.48\linewidth}
\centering
 \includegraphics[width=\textwidth]{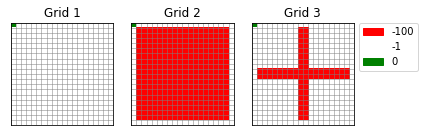}
\caption{Reward map of the three gridworlds}\label{fig:gridworlds}
\end{minipage}
\hspace{0.5cm}
\begin{minipage}[b]{0.48\linewidth}
\centering
\includegraphics[width=\textwidth]{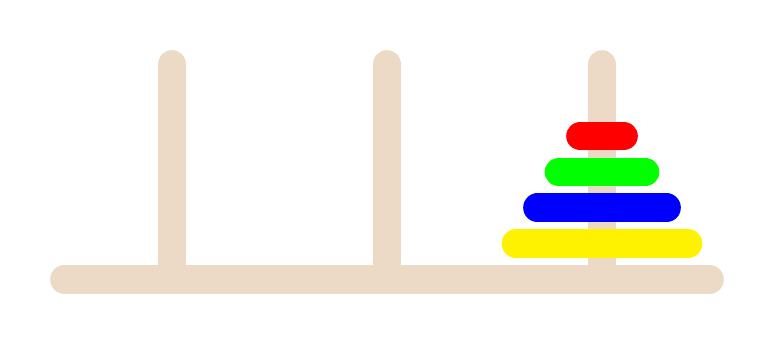}
   \caption{Final state in the Hanoi Towers experiments}\label{fig:hanoi_final}
\end{minipage}
\end{figure}

%

\paragraph{Gridworlds}

We will first test our approach on three different 20 $\times$ 20 gridworlds with different reward maps, as it can be seen in Figure \ref{fig:gridworlds}. The pool of demonstrations in these gridworlds consists of 100 expert demonstrations which result from the expert policy starting from 100 different states in the grid. \ifthenelse{\boolean{reftoappendix}}{The curriculums used for these environments can be viewed in Appendix \ref{appendix:curriculums}.}{}

\paragraph{Hanoi Towers}

To study the robustness of our approach with respect to domains with sparse rewards, we consider the puzzle of Hanoi Towers, whose goal is to move some rings (here 4) lying around different rods (here 3) to a final configuration (see Figure \ref{fig:hanoi_final}). In this puzzle, each move consists in taking the disk which is at the top of a stack, with the rule that large disks always have to be under smaller disks. We consider a batch of expert demonstrations, corresponding to the demonstration of an optimal teacher solving the puzzle for each of the possible starting states (which makes $3^4 = 81$ demonstrations).

\begin{figure}[ht]
\begin{minipage}[b]{0.48\linewidth}
\centering
\includegraphics[width=\textwidth]{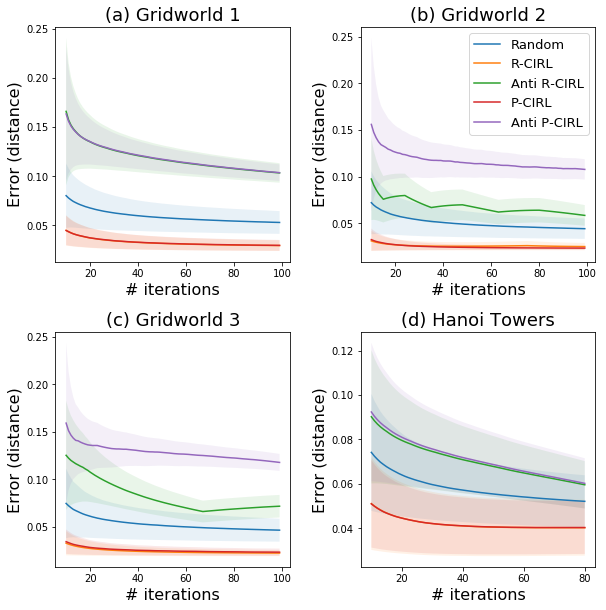}
\caption{CIRL: Evolution of the error}\label{fig:cirl_error}
\end{minipage}
\hspace{0.5cm}
\begin{minipage}[b]{0.48\linewidth}
\centering
\includegraphics[width=\textwidth]{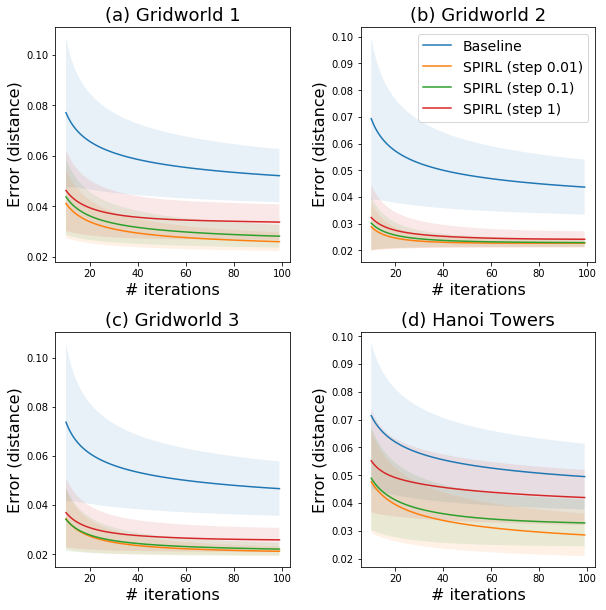}
\caption{SPIRL: Evolution of the error}\label{fig:spirl_error}
\end{minipage}
\end{figure}

%

\paragraph{Results (CIRL)}

In Figure \ref{fig:cirl_error}, we show the evolution of the features expectation matching error. Table \ref{tab:cirl_final} reports the error at the end of the training process (when every expert demonstration has been given once).

\begin{table}
 \centering
\begin{tabular}{c|c|c|c|}
\cline{2-4}
& Random   & R-CIRL            & P-CIRL            \\ \hline
\multicolumn{1}{|c|}{Grid 1} & 5.3 $\pm$ 1.2 & \textbf{2.9 $\pm$ 0.5} & 3 $\pm$ 0.6           \\ \hline
\multicolumn{1}{|c|}{Grid 2} & 4.4 $\pm$ 1.1 & 2.5 $\pm$ 04          & \textbf{2.3 $\pm$ 0.2} \\ \hline
\multicolumn{1}{|c|}{Grid 3} & 4.7 $\pm$ 1.2 & \textbf{2.2 $\pm$ 0.3} & 2.3 $\pm$ 04          \\ \hline
\multicolumn{1}{|c|}{Hanoi}  & 5.2 $\pm$ 1.2 & \textbf{4 $\pm$ 1.3}  & \textbf{4 $\pm$ 1.2}  \\ \hline
\end{tabular}
\caption{CIRL: Error  $(\times 10^{-2})$ at the end of the training}
\label{tab:cirl_final}
\end{table}

\paragraph{Results (SPIRL)}

In Figure \ref{fig:spirl_error}, we show the evolution of the features expectation matching error. Table \ref{tab:spirl_final} reports the error at the end of the training process (when every expert demonstration has been given once). The different curves correspond to different update steps $\lambda$ in the SPIRL algorithm.

\begin{table}
\centering
\begin{tabular}{c|c|c|c|c|}
\cline{2-5}
& Baseline & $\lambda = 0.01$ & $\lambda = 0.1$ & $\lambda = 1$ \\ \hline
\multicolumn{1}{|c|}{Grid 1} & 5.2 $\pm$ 1.1   & \textbf{2.6 $\pm$ 0.4}    & 2.8 $\pm$ 0.5            & 3.4 $\pm$ 0.7          \\ \hline
\multicolumn{1}{|c|}{Grid 2} & 4.4 $\pm$ 1     & \textbf{2.3 $\pm$ 0.2}    & \textbf{2.3 $\pm$ 0.2}   & 2.4 $\pm$ 0.3          \\ \hline
\multicolumn{1}{|c|}{Grid 3} & 4.7 $\pm$ 1.1   & \textbf{2.1 $\pm$ 0.2}    & 2.2 $\pm$ 0.3            & 2.6 $\pm$ 0.5          \\ \hline
\multicolumn{1}{|c|}{Hanoi}  & 5 $\pm$ 1.2       & \textbf{2.8 $\pm$ 0.8}      & 3.3 $\pm$ 0.8              & 4 $\pm$ 1              \\ \hline
\end{tabular}
\caption{SPIRL: Error  $(\times 10^{-2})$ at the end of the training}
\label{tab:spirl_final}
\end{table}

\subsection{Additional Experiments}

We also study the robustness of our approach in different contexts. To do this, we will only run our experiments on the last gridworld (the right one in Figure \ref{fig:gridworlds}).

\begin{wraptable}{r}{0.6\textwidth}
  \begin{minipage}[b]{0.58\textwidth}
    \centering
\begin{tabular}{|c|c|c|}
\hline
Random & R-CIRL & P-CIRL \\ \hline
3.7 $\pm$ 0.5 & \textbf{3 $\pm$ 0.3} & 3 $\pm$ 0.3 \\ \hline
\end{tabular}
\caption{CIRL: Error (reward) at the end of the training}
\label{tab:cirl_value_final}
  \end{minipage}
 
  \begin{minipage}[b]{0.58\textwidth}
    \centering
\begin{tabular}{|c|c|c|c|}
\hline
Baseline & $\lambda = 0.01$ & $\lambda = 0.1$ & $\lambda = 1$ \\ \hline
3.8 $\pm$ 0.4 & \textbf{2.9 $\pm$ 0.2} & 3 $\pm$ 0.26 & 3.2 $\pm$ 0.3 \\ \hline
\end{tabular}
\caption{SPIRL: Error (reward) at the end of the training}
\label{tab:spirl_value_final}
  \end{minipage}
\end{wraptable}

\paragraph{Another Error Metric} 

We study our algorithms with another error metric, namely the difference in expected reward between the learner's and the teacher's policy. Results of this study are shown in Tables \ref{tab:cirl_value_final} and \ref{tab:spirl_value_final}.


%
%
%


\begin{wraptable}{r}{0.6\textwidth}
  \begin{minipage}[b]{0.58\textwidth}
    \centering
\begin{tabular}{c|c|c|c|}
\cline{2-4}
 & Random & R-CIRL & P-CIRL \\ \hline
\multicolumn{1}{|c|}{0 \%} & 4.7 $\pm$ 1.2 & \textbf{2.2 $\pm$ 0.3} & 2.3 $\pm$ 04 \\ \hline 
\multicolumn{1}{|c|}{10 \%} & 5.6 $\pm$ 1.3 & \textbf{2.3 $\pm$ 0.4} & 2.4 $\pm$ 0.4 \\ \hline
\multicolumn{1}{|c|}{30 \%} & 8.9 $\pm$ 1.6 & \textbf{2.5 $\pm$ 0.5} & \textbf{2.5 $\pm$ 0.5} \\ \hline
\end{tabular}
\caption{CIRL: Error $(\times 10^{-2})$ with suboptimal demonstrations}
\label{tab:cirl_suboptimal}
  \end{minipage}
 
  \begin{minipage}[b]{0.58\textwidth}
    \centering
\begin{tabular}{c|c|c|c|c|}
\cline{2-5}
 & Baseline & $\lambda = 0.01$ & $\lambda = 0.1$ & $\lambda = 1$ \\ \hline
\multicolumn{1}{|c|}{0 \%} & 4.7 $\pm$ 1.1   & \textbf{2.1 $\pm$ 0.2}    & 2.2 $\pm$ 0.3            & 2.6 $\pm$ 0.5          \\ \hline
\multicolumn{1}{|c|}{10 \%} & 4.8 $\pm$ 1 & \textbf{2.1 $\pm$ 0.2} & 2.2 $\pm$ 0.3 & 2.5 $\pm$ 0.5 \\ \hline
\multicolumn{1}{|c|}{30 \%} & 14 $\pm$ 3.4 & \textbf{6.3 $\pm$ 0.5} & 6.4 $\pm$ 0.7 & 7.5 $\pm$ 1.4 \\ \hline
\end{tabular}
\caption{SPIRL: Error $(\times 10^{-2})$ with suboptimal demonstrations}
\label{tab:spirl_suboptimal}
  \end{minipage}
\end{wraptable}

\paragraph{Suboptimal Demonstrations} 

We study the influence of suboptimal demonstrations on the learner's performance. The teacher provides demonstrations following the optimal policy, but with a fixed proportion of random actions, respectively 0 \% (results from Tables \ref{tab:cirl_value_final} and \ref{tab:spirl_value_final}), 10 \% and 30 \%. Results are presented in Tables \ref{tab:cirl_suboptimal} and \ref{tab:spirl_suboptimal}. The error is the difference fetween the learner's policy and the teacher's optimal policy (i.e with no random actions).

%

\begin{figure}[ht]
\begin{minipage}[b]{0.48\linewidth}
\centering
\includegraphics[width=\textwidth]{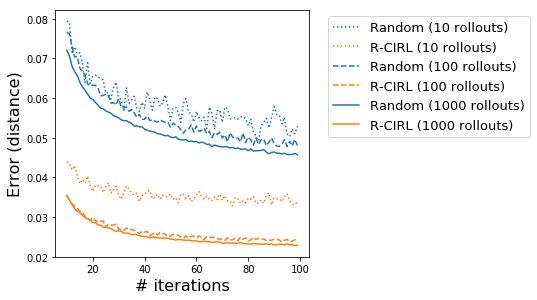}
\caption{CIRL with Monte-Carlo estimation}\label{fig:cirl_mc}
\end{minipage}
\hspace{0.5cm}
\begin{minipage}[b]{0.48\linewidth}
\centering
\includegraphics[width=\textwidth]{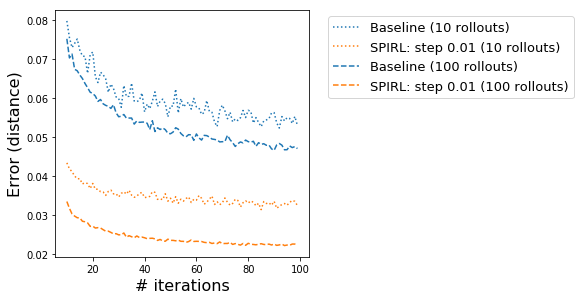}
  \caption{SPIRL with Monte-Carlo estimation}\label{fig:spirl_mc}
\end{minipage}
\end{figure}

\paragraph{Computation of Features Expectations} So far, to update her belief about the environment, the MaxEnt IRL learner relies on the exact computation of her features expectations vector (i.e $\mu_{\pi_{w_t}}$ in Algorithms \ref{algo:onlinemaxent} and \ref{algo:spirl}) using Soft Value Iteration, which is often intractable in higher dimensional or continuous control environments. This is why we study whether our algorithms outperform the baselines if the learner's features expectations vector is computed using a finite number of trajectories (Monte-Carlo rollouts). To study the robustness, we add stochasticity in the environment's transition dynamics (here 20 \% of the learner's and teacher's moves will be random). Results of this experiment are presented in Figures \ref{fig:cirl_mc} and \ref{fig:spirl_mc}.

%


\section{Conclusions}\label{sec:conclusions}

In this paper, we presented CIRL and SPIRL, two approaches guided by theoretical insights of the MaxEnt IRL theory, which deal with the absence of teacher-learner interaction in Inverse Reinforcement Learning. Using experiments in simulated environments and with a real robotic task, we show that our algorithms allow an improvement in the learning speed compared to a teacher providing demonstrations at random (for CIRL) and to a learner processing a whole batch of demonstrations (for CIRL). Our experimental results show that CIRL will provide first, in the gridworlds, demonstrations which are close to a goal state, without specifying any obstacles or dangerous zones. Future work could focus on designing curriculums for IRL with hard constraints such as the fact that obstacles should be learned as fast as possible in the learning process.

\acknowledgments{This work has received funding in part from the European Research Council (ERC) under the European Union's Horizon 2020 research and innovation program (grant agreement n 725594 - time-data), and the Swiss National Science Foundation (SNSF) under grant number 407540\_167319.}


\bibliography{cirl}

\ifthenelse{\boolean{showappendix}}{
\clearpage
\appendix

\section{Maximum Entropy Inverse Reinforcement Learning}
\label{appendix:maxentirl}

In the Maximum Causal Entropy framework, we assume a probability distribution over the trajectories based on the exponential of the reward of these trajectories. We have
\[
\begin{split}
\mathbb{P}(\xi | w) & \propto \exp {\left\langle w, \mu_\xi\right\rangle} ~=~ \exp{\left\langle w, \sum_{t=0}^{\infty} \gamma^t \phi(s_t, a_t)\right\rangle} \\
&~=~ \exp\left( \gamma^t \sum_{t=0}^{\infty} \langle w, \phi(s_t, a_t)\rangle \right).
\end{split}
\]
To satisfy this condition, the action probability $\pi_w$ is given by 
\[
\forall (s,a) \in \mathcal{S} \times \mathcal{A},~\pi_w(s, a) ~=~ \exp\left(Q_w(s,a) - V_w(s)\right),
\]
where
$$\begin{cases}
Q_w(s,a) ~=~ \langle w, \phi(s, a)\rangle + \gamma \sum_{s'} T(s'~|~s,a) V_w(s') \\
V_w(s) ~=~ \log \left( \sum_{a} \exp{Q_w(s,a)}\right) .
\end{cases}$$

Let us define the soft optimal Bellman operator $\mathcal{B}_{w} : \mathbb{R}^{|\mathcal{S}|} \to \mathbb{R}^{|\mathcal{S}|}$ such that $\forall s \in \mathcal{S}$,
$$\mathcal{B}_{w}(V)(s) = \log \underset{a\in \mathcal{A}}{\sum} \exp (\langle w, \phi(s, a)\rangle  + \gamma \underset{s'\in \mathcal{S}}{\sum} T(s'|s,a) V(s')).$$

We can prove that $\mathcal{B}_{w}$ is a contraction and use the fact that $V_w$ is a fixed point of $\mathcal{B}_{w}$ to compute $V_w$. This procedure called ``Soft Value Iteration'' is described in Algorithm~\ref{algo:softvalueiteration}.

\begin{algorithm}[H]
\caption{Soft Value Iteration}\label{algo:softvalueiteration}
\begin{algorithmic}[1]
\Require Weights vector $w$
\State Initialize $V \in \mathbb{R}^{|\mathcal{S}|}$
\While{not converged}
\For{$(s,a) \in \mathcal{S} \times \mathcal{A}$}
\State $Q(s,a) \leftarrow \langle w, \phi(s, a)\rangle + \gamma \underset{s'\in \mathcal{S}}{\sum} T(s'|s,a) V(s')$
\EndFor
\For{$s \in \mathcal{S}$}
\State $V(s) \leftarrow \log \underset{a\in \mathcal{A}}{\sum} \exp Q(s,a)$
\EndFor
\EndWhile
\State \Return $\pi_w = \left(\exp\left(Q(s,a) - V(s)\right)\right)_{s \in \mathcal{S}, a \in \mathcal{A}}$
\end{algorithmic}
\end{algorithm}

\section{Visualization of the Curriculums}
\label{appendix:curriculums}

Since there is a one-to-one mapping between the expert demonstrations and the cells of the gridworld, we can vizualize the order of the demonstrations in the curriculums in the map. Based on the considerations made in Section \ref{subsection:CIRL}, the curriculums which are chosen for each environment are shown in Figure \ref{fig:curriculums}. 

The bright areas correspond to the starting state of the demonstrations with high reward for P-CIRL or high probability for P-CIRL (i.e. demonstrations which will be given early in the learning process).

\begin{figure}[!htb]
\begin{center}
\includegraphics[width=0.5\linewidth]{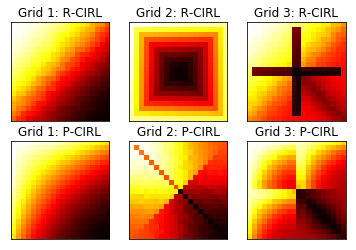}
\end{center}
\caption{Vizualisation of the curriculums in the different environments}
\label{fig:curriculums}
\end{figure}

\section{Robotics Experiment}
\label{appendix:robotics}

\paragraph{Experimental Setup} 

In this section, we  present an application of CIRL and SPIRL for the learning of a robotic task. The task consists of moving a ball held with a spoon and dropping the ball in a cup, while avoiding obstacles. The robot used for this experiment is a Franka Emika Panda robot. The experimental setup can be seen in Figure \ref{fig:setup_robot}. The state space of the robot is of 7 dimensions, corresponding to its joint angles. We use a Gaussian Mixture Model to split the continuous state space into a discrete state space (with 40 states). The environment is then a Markov Decision Process with 40 states and 40 actions (which correspond to move to each of the states). The states features correspond to a one-hot encoding of the 40 states. The expert reward will be computed using a distance between the different states of the GMM. We record by kinesthetic teaching 33 different demonstrations of the task, with different initial states. The movements required to reach the cup while keep the spoon horizontal are varied, notably due to the robot's joint angle limits and the obstacle.

To apply the framework of discrete Markov Decision Processes to this continuous environment, in a first phase, we record with the robotic arm many different states in which the robot can be (without trying to record this task). Once these states are recorded, we augment them with the 33 demonstrations recorded for the task.

Then, we use a Gaussian Mixture Model to segment all the states recorded during the two first phases into 40 clusters. This discretization of the state space allows a more compact representation than discretizing each coordinate. The modeling of the state space with multivariate normal distributions provides a convenient way to move from a state to another using classical robotics techniques such as linear quadratic regulator (LQR).

Note that since we have the constraint that the spoon should be kept approximately horizontal so that the ball does not fall, the dimensionality of the allowed joints configurations is reduced. 

To make it more realistic, we restrict actions such that we can only move from a state to adjacent states. To compute the distance between two states, we compute the integral of the product of the probability density functions of the states (seen as multivariate normal distributions). The historical lowest value of this integral (computed over the data recorded in the two first phases) will serve as a threshold to determine which states are adjacent and which are not, even if they were not visited during the recording phase.

Instead of setting hand-engineered reward weights for the teacher, we compute the reward based on the integral of the product of the probability density functions of the states.

\begin{figure}[!htbp]
\begin{center}
\includegraphics[width=0.6\columnwidth]{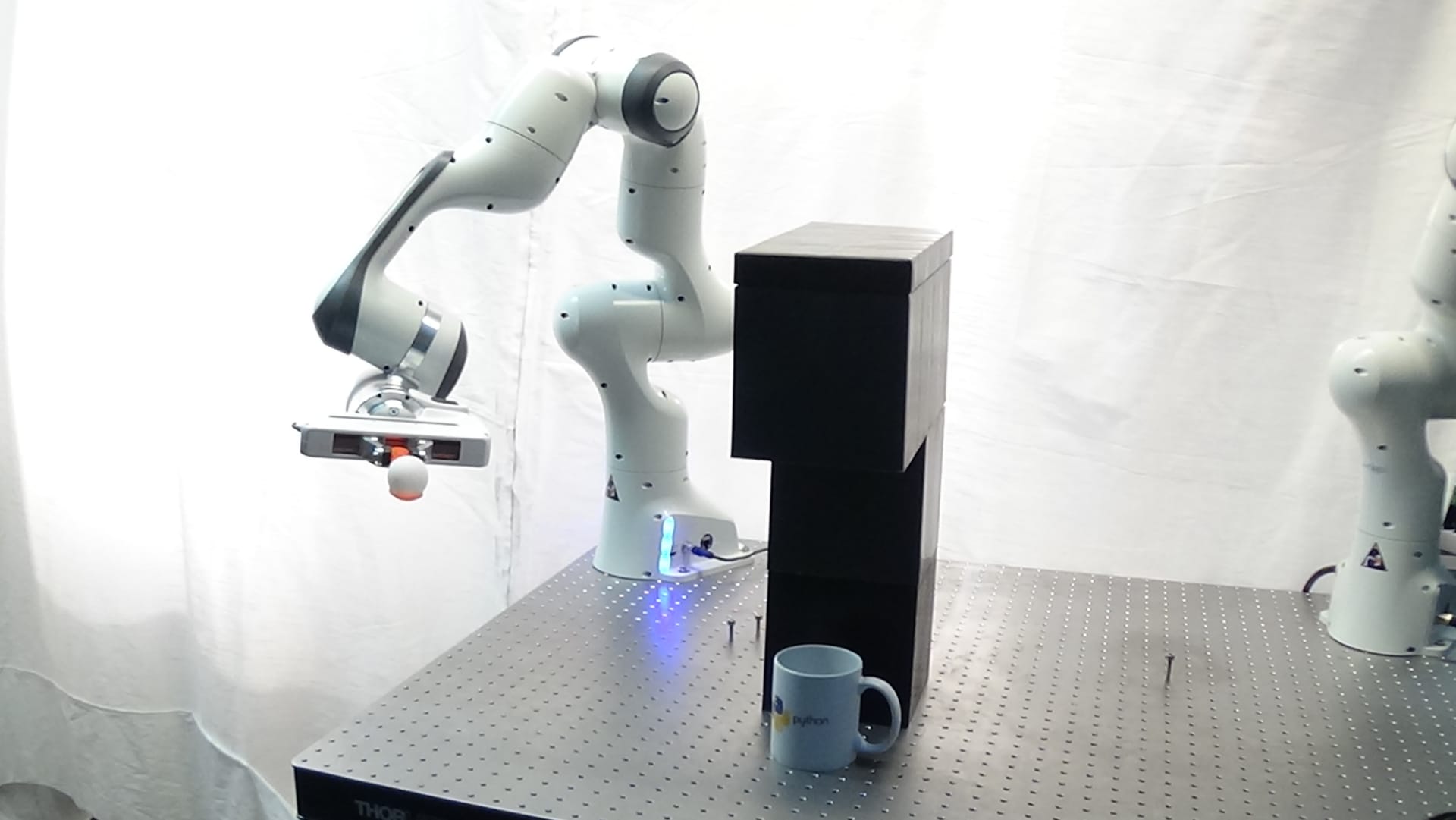}
\end{center}
\caption{Experimental setup for the robot manipulation task}
\label{fig:setup_robot}
\end{figure}

\paragraph{Results} In this experiment, we study CIRL with the reward-based curriculum. In addition to the error in features expectations between the set of expert demonstrations and the learner's policy, we look at the expected reward of a trajectory (using the distance between the states of the GMM). Results of this experiment are presented in Figure \ref{fig:robotics_results}. 

\begin{figure}[!htbp]
\begin{center}
  \includegraphics[width=0.6\columnwidth]{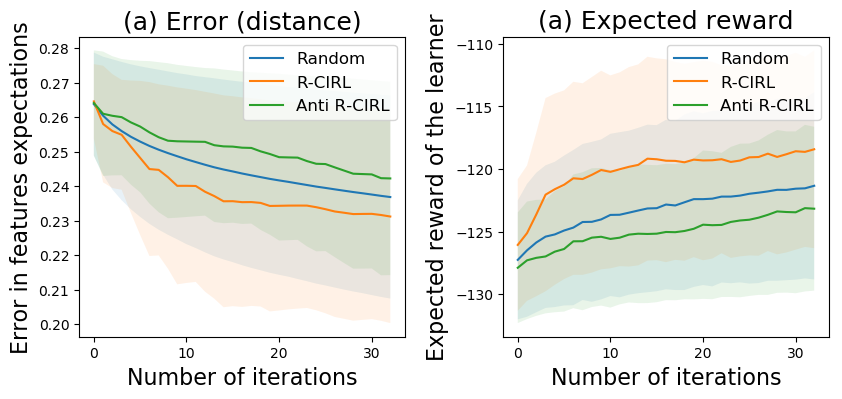}
  \end{center}
  \caption{Robotics experiment: results}\label{fig:robotics_results}
\end{figure}

\paragraph{Comments on the Video} In the video attached to this paper, we present the experimental setup and show examples of learned policies in the robotics experiment. 

The first sequence of the video presents a human demonstrator showing the robot how to do the task. 

The three following sequences show three expert demonstrations: the first and the last demonstration of the reward-based curriculum and another demonstration. We replay these demonstrations without the expert for a better visualization.

The two last sequences of the video present two examples of trajectories produced by policies learned with MaxEnt IRL. The first one comes from a policy learned after one step of learning with the curriculum teacher, the second one after one step of learning with the anticurriculum teacher. We see that the first trajectories goes directly to the goal whereas the second one seems to explore a lot of states before reaching the goal.

\subsection{Discussion}

Based on the results presented in this section, CIRL outperforms the random teacher in most of the environments. The two curriculums designed have comparable performance. SPIRL also outperforms the batch learner, with its performance depending on the update step $\lambda$. 

SPIRL and CIRL show robustness with respect to stochastic transitions and inexact computation of the features expectations, which is a first step towards adapting these strategies to continuous control tasks. In \cite{levine2012continuous}, the authors propose a gradient update rule for continuous Inverse Optimal Control (based on an approximation of the soft Bellman policy using a second-order Taylor expansion of the reward). Future work could focus on extending our analysis for this new update rule and devise new teaching strategies 

Also, our algorithms do not outperform the baselines when the learner is using an adaptive optimization strategy. Future work will investigate whether update rules  better suited for online learning could aleviate this issue, e.g. reusing the previously used demonstrations to avoid forgetting.

In the robotics experiment, even if CIRL outperforms the random teacher using the different metrics, the difference with the random teacher is not as significant as in the other environments. This might be due to the modeling of the environment into an MDP and to the fact that the human demonstrator is not optimizing a reward function in a Markov Decision Process but rather doing the task on intuition, which might lead to sub-optimal demonstrations in the batch.

}{}

\end{document}